\documentclass[11pt]{article}

\usepackage[preprint]{acl}

\usepackage{times}
\usepackage{hyperref}
\usepackage[raster,skins, most]{tcolorbox} %
\usepackage{multirow}   
\usepackage[table]{xcolor} 
\definecolor{lightblue}{rgb}{0.9, 0.95, 1.0} 
\usepackage{url}
\usepackage{latexsym}
\usepackage{algorithm}
\usepackage{algpseudocode}
\usepackage{amsmath}
\usepackage{amssymb}
\usepackage{booktabs}
\usepackage[T1]{fontenc}

\usepackage[utf8]{inputenc}

\usepackage{microtype}

\usepackage{inconsolata}

\usepackage{graphicx}

\title{WebClipper: Efficient Evolution of Web Agents with Graph-based Trajectory Pruning}

\author{
 \textbf{Junjie Wang},
 \textbf{Zequn Xie},
 \textbf{Dan Yang},
 \textbf{Jie Feng},
 \textbf{Yue Shen},
 \textbf{Duolin Sun},
\\
 \textbf{Meixiu Long},
 \textbf{Yihan Jiao},
 \textbf{Zhehao Tan},
 \textbf{Jian Wang},
 \textbf{Peng Wei},
 \textbf{Jinjie Gu}\thanks{Corresponding author}
\\
 Ant Group
\\
 \small{
   \textbf{Correspondence:} {wjj417805@antgroup.com,wangjj2018@zju.edu.cn,jinjie.gujj@antgroup.com}
 }
}

\begin{document}
\maketitle
\begin{abstract}
Deep Research systems based on web agents have shown strong potential in solving complex information-seeking tasks, yet their search efficiency remains underexplored. We observe that many state-of-the-art open-source web agents rely on long tool-call trajectories with cyclic reasoning loops and exploration of unproductive branches. To address this, we propose WebClipper, a framework that compresses web agent trajectories via graph-based pruning. Concretely, we model the agent’s search process as a state graph and cast trajectory optimization as a minimum-necessary Directed Acyclic Graph (DAG) mining problem, yielding pruned trajectories that preserve essential reasoning while eliminating redundant steps. Continued training on these refined trajectories enables the agent to evolve toward more efficient search patterns and reduces tool-call rounds by about 20\% while improving accuracy. Furthermore, we introduce a new metric called F-AE Score to measure the model's overall performance in balancing accuracy and efficiency. Experiments demonstrate that WebClipper compresses tool-call rounds under excellent performance, providing practical insight into balancing effectiveness and efficiency in web agent design. Code is available at \url{https://github.com/AQ-MedAI/AntAFu-DeepResearch}.
\end{abstract}

\section{Introduction}

With the continuous evolution of Large Language Models (LLMs), artificial intelligence systems have transformed from static text-based models into sophisticated agents capable of utilizing tools and interacting with environments \cite{kimik2,glm45,xiong2026scaling}. Among these, web agents have demonstrated remarkable capabilities in complex information-seeking, completing challenging tasks in tens of minutes that would typically require humans several hours. Representative examples include commercial systems such as OpenAI's Deep Research \cite{openaidr}, Gemini \cite{geminiresearch}, and Claude \cite{clauderesearch}, alongside emerging open-source alternatives like Tongyi-DeepResearch \cite{tongyidrs} and MiroThinker \cite{mirothinker}.

\begin{figure}
    \centering
    \includegraphics[width=\linewidth]{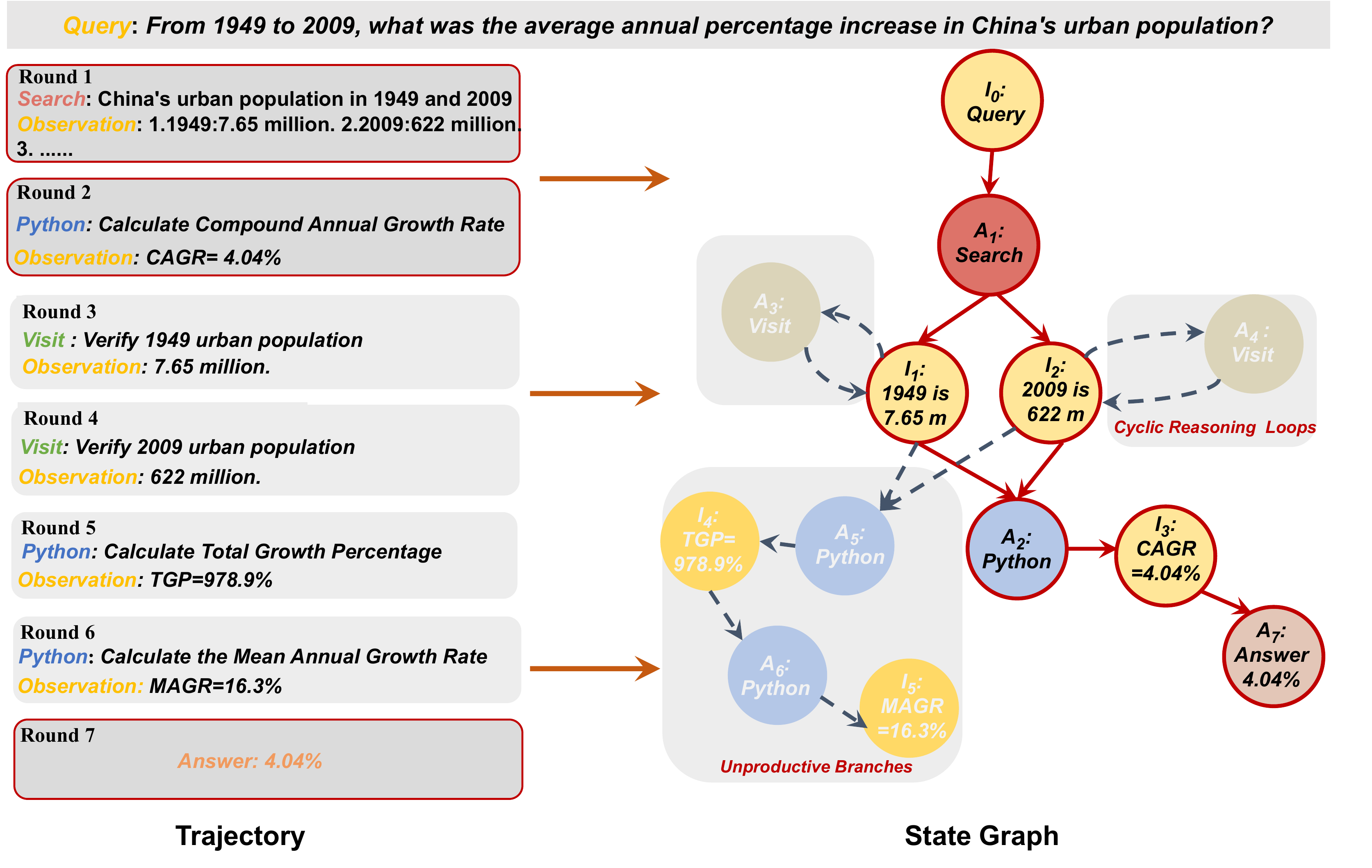}
    \caption{The trajectory of a web agent can be built as a graph. The minimum number of steps to solve the problem is the minimum necessary DAG from the Query node (I0) to the final Answer Action node (A7).}
    \label{fig:graph_example}
\end{figure}
However, current open-sourced web agents primarily focus on the final problem-solving accuracy while paying little attention to efficiency during the search process. In pursuit of higher accuracy, these agents continuously scale up search depth and context length \cite{tongyidrs}, leading to extremely long contexts and excessive tool usage. For example, Tongyi-DeepResearch uses a 128K context length and up to 100 tool-call rounds, while MiroThinker sets a maximum context length of 256K and allows up to 600 tool-call rounds. Considering the long inference time and the high costs of commercial search tools (e.g., Google Search and Jina Reader), the user experience in practice is far from ideal. 

To understand what causes such inefficiency, we conduct a deeper analysis of the agent's search behavior. Prior work \cite{lostinmaze,webleaper} highlighted that effective actions are sparsely distributed across long trajectories. For many failure cases, the agent repeatedly re-searches information it has already obtained or over-focuses on noisy signals \cite{lostinmaze}, causing it to drift away from the correct direction, which should ideally be avoided. To systematically identify such inefficiency patterns, we model the trajectory of the agent as a state graph. As is illustrated in Figure \ref{fig:graph_example}, the agent's action and environmental observation can be abstracted as nodes in the graph. This formalization reveals two major inefficient patterns: \textbf{cyclic reasoning loops} and \textbf{unproductive branches} that diverge from the correct solution, while the ideal path should be the minimum DAG from the original query to the final answer.

The above observation motivates us to prune these inefficient patterns to construct a more robust web agent. However, training a robust web agent from scratch remains both costly and challenging due to complex data synthesis pipelines and multi-stage training paradigms \cite{tongyidrs,step-drs} that range from agentic mid-training to SFT and RL. This leads us to explore a different direction: \textbf{Instead of building a new agent from scratch, can we evolve pre-existing, high-performance but low-efficiency web agents into more efficient ones by pruning their inefficient patterns?}

To achieve this, we introduce WebClipper, a novel framework designed to optimize the search behavior of web agents toward a better accuracy-efficiency balance. Specifically, our framework consists of:
1) Trajectory to State-Graph transformation: transforming raw trajectories into state graphs by abstracting agent actions and environment information.
2) Pruning via an approximate minimal necessary DAG (MNDAG): mining an approximate MNDAG that connects initial information nodes to final action nodes, thereby pruning redundant steps.
3) Coherence-aware thought rewriting: rewriting the agent's thoughts on the pruned trajectories to ensure semantic consistency and usability.
4) Agent Evolution: training existing agents to improve efficiency based on collected trajectories combined with a hybrid evolution strategy.
To quantify the accuracy-efficiency trade-off, we further propose a new evaluation metric, \textbf{F-AE Score}. Instead of separately reporting performance and resource usage, the F-AE Score reflects how well a web agent balances these two aspects, providing a direct view for comparing different optimization strategies and guiding the design of more practical web agents.

Experiments on multiple benchmarks show that WebClipper reduces tool-call rounds and token usage by about 20\% while maintaining or even improving accuracy. Our contributions are summarized as follows:

1) We propose WebClipper, a novel pruning method for existing Deep Research–style web agents, enabling them to evolve toward a more efficient search behavior.

2) Our methods explicitly target the accuracy–efficiency trade-off, together with the F-AE score as a unified metric to evaluate this balance.

3) We evaluate WebClipper on multiple benchmarks and empirically demonstrate its good balance between accuracy and efficiency. 

\begin{figure*}
    \centering
    \includegraphics[width=\linewidth]{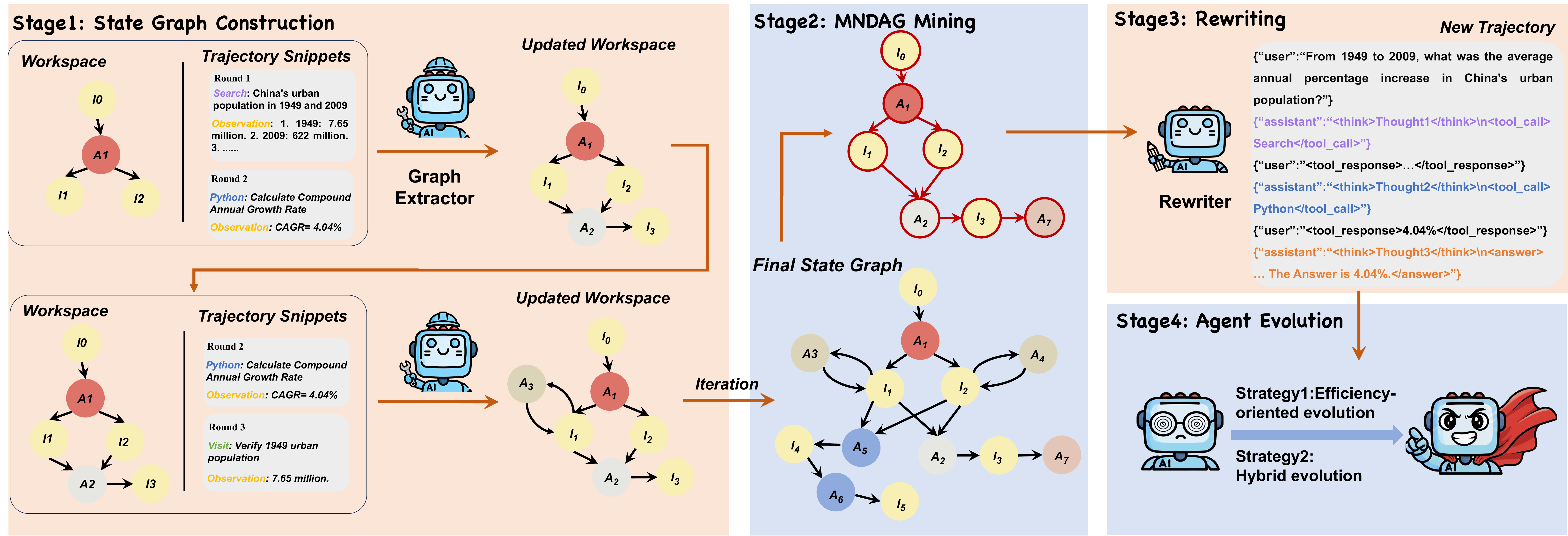}
    \caption{The overview of WebClipper}
    \label{fig:main_graph}
\end{figure*}

\section{Related Work}

\textbf{Deep Research Agents.} Methods for web agents can be broadly divided into two categories. The first is training-free approaches, which solve tasks by designing multi-agent collaborative architectures, such as OpenDeepResearch \cite{OpenDeepResearch}, GPT Researcher \cite{GPTResearch}, and WebWeaver \cite{webweaver}. These works typically focus on how to structure the agent state space, using context engineering to compress and share context across agents so that they perform better on long-horizon, complex tasks. The second category is training-based approaches, which aim to train a single, powerful core agent that can flexibly use various tools within a constructed environment. To obtain such agents, a large body of work focuses on synthesizing training data for web agents, generating complex multi-hop questions from open webpages or knowledge graphs \cite{websailor,webshaper,lpkg}, and then applying SFT or RL to improve the agent’s capability on challenging tasks \cite{webexplorer,webthinker, ReSearch}. However, these methods almost exclusively target end-to-end task success rates, while paying very little attention to the efficiency of web agents.

\noindent \textbf{Efficient Reasoning in LLMs.} With the emergence of reasoning models such as OpenAI-o1 \cite{o1} and DeepSeek-R1 \cite{r1}, there has been growing interest in efficient reasoning for single LLMs. A simple yet effective line of work is prompt-based, where explicit instructions are added to the prompt to encourage the model to reason in a more efficient manner \cite{cot-prompt1,cot-prompt2,cot-prompt3}. Beyond prompting, many methods rely on training-based strategies: for example, compressing the long chain-of-thoughts (CoT) into shorter ones to train a model that acquires short-thinking capabilities and maintains performance under low-resource settings \cite{cot-train1,cot-train2,cot-train3}; or incorporating length-related rewards into RL training so that the model learns to discover more efficient reasoning paths \cite{cot-rl1,cot-rl2,cot-rl3}. These compression techniques for single models inspire our design of methods to improve the search efficiency of web agents.

\section{Methodology}
In this section, we present \textbf{WebClipper}, a framework to evolve an existing Deep Research–style web agent into a more efficient one. As shown in Figure \ref{fig:main_graph}, our framework consists of four main components:  (1) constructing state graphs from raw trajectories, (2) mining an approximate MNDAG for pruning, (3) coherence-aware thought rewriting, followed by (4) agent evolution based on the pruned trajectories.

\subsection{Preliminaries and Notation}
Let a query be denoted by \(q\). Given \(q\), a web agent interacts with the environment through a trajectory:
\[
\tau = \bigl( o_0, r_1, a_1, o_1, \dots, r_T, a_T \bigr),
\]
where \(o_t\) is the observation from the environment at round \(k\) (with \(o_0=q\)), \(r_t\) is the agent’s thought, and \(a_t\) is the agent’s action. Actions include tool invocations (e.g., \textsc{Search}, \textsc{Visit}, \textsc{Python}) and the final answer \(\textsc{Answer}\). 
Our goal is to transform each raw trajectory \(\tau\) samples from raw agent $\mathcal{M}$ into an accurate and efficient trajectory \(\tilde{\tau}\), and then use a collection of such trajectories to train a model $\mathcal{M}'$ that achieves comparable accuracy with fewer action steps.

\subsection{Initial Trajectory Collection and Filtering}
\label{rejection_sample}
We first collect question–answer (QA) pairs from public datasets such as WebShaper \cite{webshaper}, WebDancer \cite{webdancer}, WebExplorer \cite{webexplorer}, TaskCraft \cite{taskcraft}, and Voyager \cite{mirothinker}. Using the pre-built environment, we distill trajectories from the existing web agent $\mathcal{M}$, which follows a ReAct-style \cite{yao2023react} loop of observation–think–action.

For each \(q\), we first sample $K$ distinct trajectories \(\{ \tau^{(k)} \}_{k=1}^K\) from $\mathcal{M}$. We then employ a rejection sampling strategy: let \(\mathrm{PR}(q)\in[0,1]\) denote the pass rate of \(q\) on the target task, we retain only trajectories of queries satisfying:
$0 < \mathrm{PR}(q) \le 0.5,$
which keeps the task challenging. These trajectories constitute the input to our pruning pipeline.

\subsection{From Trajectory to State Graph}

\subsubsection{State Graph Definition}
Given a trajectory \(\tau\), we construct a directed graph
$\mathcal{G} = (\mathcal{V}^A \cup \mathcal{V}^I, \mathcal{E}),$
where \(\mathcal{V}^A = \{ A_1, \dots, A_T \}\) is the set of \textbf{Action nodes}. Each \(A_t\) abstracts the agent’s thought and action at step \(t\).
\(\mathcal{V}^I = \{ I_0, I_1, \dots \}\) is the set of \textbf{Information nodes}, representing atomic pieces of information obtained from the environment, including the initial query. 

We denote the initial query node as \(I_0\), and the final answer node as \(A_T\). Edges \(\mathcal{E}\) capture the dependency between actions and information:
\(I \rightarrow A\) if action \(A\) is taken based on information \(I\);
\(A \rightarrow I\) if information \(I\) is produced as a result of action \(A\). This yields a bipartite, directed structure between \(\mathcal{V}^A\) and \(\mathcal{V}^I\).

\subsubsection{State Graph Construction}
We construct \(\mathcal{G}\) from \(\tau\) with an LLM-based extractor. First, for each step \(t\) with internal thought \(r_t\) and action \(a_t\), the extractor summarizes \((r_t, a_t)\) into a compact Action node \(A_t\) (recording action type and goal), yielding \(\{A_t\}_{t=1}^T\). We then build information nodes \(\mathcal{V}^I\) and edges \(\mathcal{E}\) iteratively using a workspace \(\mathcal{W}\) that stores current information nodes and links. Initially, \(\mathcal{W} = \{ I_0 \}\), where \(I_0\) encodes the original query. For each step \(t = 0, \dots, T-1\), we feed the snippet \((A_t, o_t, A_{t+1})\) and \(\mathcal{W}\) to the extractor, prompting it to:

1) \textbf{Decompose observation into atomic information.}
   \(o_t\) is decomposed into atomic units \(\{I^*\}\). Each \(I^*\) is matched against existing nodes in \(\mathcal{W}\); on a semantic match, we add \(A_t \rightarrow I\); otherwise we create a new information node \(I^*\), insert it into \(\mathcal{V}^I\) and \(\mathcal{W}\), and add \(A_t \rightarrow I^*\).

2) \textbf{Link new action to supporting information.} 
   The extractor analyzes $A_{t+1}$ to identify a set of information nodes \(\mathcal{S}_{k} \subseteq \mathcal{V}^I\) in \(\mathcal{W}\) that the agent relies on when executing \(A_{t+1}\). For each \(I \in \mathcal{S}_{k}\), we add an edge \(I \rightarrow A_{t+1}\).

This process continues until the final answer action \(A_T\) is reached. The result is a state graph \(\mathcal{G}\) that explicitly encodes the dependency between all actions and information along the trajectory.

\subsection{Pruning via MNDAG}

Given the state graph \(\mathcal{G}\), we aim to identify the approximate minimal subgraph that is necessary and sufficient to support the final answer. Intuitively, actions that do not contribute any information used (directly or indirectly) by the answer are deemed redundant and should be pruned.

We treat the initial query node \(I_0\) as the source and the final answer node \(A_T\) as the sink. Each action node \(A_t\) is assigned a unit cost \(c(A_t) = 1\), and each information node cost is set to zero, i.e., \(c(I_t) = 0\). Our objective is to find an approximate minimal-cost directed acyclic subgraph \(\mathcal{G}^\star\) that connects \(I_0\) to \(A_T\) and preserves all necessary dependencies.

We approximate this by:

1) \textbf{Shortest-path forward search.} 
   We run a Dijkstra-style shortest-path algorithm on \(\mathcal{G}\) from \(I_0\) to \(A_T\), using node costs \(c(\cdot)\) aggregated along the path. This yields the shortest path
   $P = (I_0 \rightarrow \cdots \rightarrow A_T)$,
   which captures one minimal-cost path from query to answer.

2) \textbf{Backward closure of necessary predecessors.}  
   Starting from \(A_T\), we perform a reverse traversal on \(\mathcal{G}\), recursively adding predecessor nodes that are on some shortest path contributing to the answer. This ensures that we do not miss necessary branching dependencies. The resulting set of nodes \(\mathcal{V}^\star \subseteq \mathcal{V}^A \cup \mathcal{V}^I\) and edges \(\mathcal{E}^\star\) form a MNDAG:
   $\mathcal{G}^\star = (\mathcal{V}^\star, \mathcal{E}^\star).$ A detailed algorithm description of the MNDAG is expanded at Algorithm \ref{alg:mndag} in Appendix \ref{sec:method_details}.

All action nodes \(A_t \notin \mathcal{V}^\star\) are considered redundant and will be removed from the trajectory, thus obtaining a necessary actions set $\mathcal{A}^{\star}$. To improve robustness, we repeat the graph construction and MNDAG mining process three times for the same raw trajectory \(\tau\), obtaining three candidate sets of necessary actions:
$\mathcal{A}^{\star(1)},\ \mathcal{A}^{\star(2)},\ \mathcal{A}^{\star(3)}.$
We then perform a majority vote at the action set level. The final set of necessary actions, $\mathcal{A}^{\star}_{final}$, is determined only if at least two of the three candidate sets are identical.

\subsection{Coherence-aware Thought Rewriting}

Directly removing intermediate steps from a trajectory may break the coherence of the ReAct loop. We therefore perform coherence-aware rewriting over the pruned trajectory via context-aware selective rewriting and perplexity-based selection.

Given $\mathcal{A}^{\star}_{final}$, we map it back to a pruned trajectory by retrieving each selected thought–action pair and its following observation from \(\tau\), yielding
\[
\tilde{\tau} = \bigl( o^{\text{new}}_0, r^{\text{new}}_1, a^{\text{new}}_1, o^{\text{new}}_1, \dots, r^{\text{new}}_L, a^{\text{new}}_L \bigr),
\]
where \(L \le T\) and all actions \(a^{\text{new}}_t\) and thought \(r^{\text{new}}_t\) correspond to nodes in $\mathcal{A}^{\star}_{final}$.

1) \textbf{Context-aware selective rewriting.} For consecutive snippets \(\bigl(r^{\text{new}}_t, a^{\text{new}}_t, o^{\text{new}}_t, r^{\text{new}}_{t+1}, a^{\text{new}}_{t+1}\bigr)\), if \(a^{\text{new}}_t\) and \(a^{\text{new}}_{t+1}\) were adjacent in the original trajectory, we keep them unchanged. Otherwise, we rewrite \(r^{\text{new}}_{t+1}\) with a rewriter LLM based on the full context, including the pruned intermediate steps, prompting the rewriter to maintain logical continuity and remove references to pruned observations in the \(r^{\text{new}}_{t+1}\), obtaining the rewritten thought \(\hat{r}^{\text{new}}_{t+1}\).

2) \textbf{Perplexity-based selection.} To align the rewritten thoughts \(\hat{r}^{\text{new}}_{t+1}\) with the base model's intrinsic reasoning style, we generate three candidate rewrites and select the one with the lowest perplexity (PPL) as calculated by the base model $\mathcal{M}$ itself. This process ensures alignment with the model's intrinsic reasoning style as much as possible. Finally, we obtain a set of high-quality pruned trajectories $\mathcal{D}_{pruned} = \{\tilde{\tau}\}$.

\subsection{Agent Evolution via Efficient and Hybrid Training} \label{Evolution}

After obtaining $\mathcal{D}_{pruned} = \{\tilde{\tau}\}$, we use them to further train the base $\mathcal{M}$, evolving it into more efficient search behavior.

We propose two evolution paradigms:

1) \textbf{Efficiency-oriented evolution}: Fine-tune $\mathcal{M}$ solely on $\mathcal{D}_{pruned}$ to maximize search efficiency:
$$\mathcal{L}_{eff} = -\sum_{\tilde{\tau} \in \mathcal{D}_{pruned}} \log P_{\mathcal{M}}(\tilde{\tau})$$

2) \textbf{Hybrid evolution}: To balance efficiency and accuracy, we construct a hybrid dataset $\mathcal{D}_{hybrid} = \mathcal{D}_{pruned} \cup \mathcal{D}_{unpruned}$, where $\mathcal{D}_{unpruned}$ contains unpruned trajectories with different queries (non-overlapping with $\mathcal{D}_{pruned}$) and similar difficulty ($0 < \mathrm{PR}(q) \leq 0.5$). Trajectories in $\mathcal{D}_{unpruned}$ are those where our MNDAG extraction finds no redundant rounds to prune. They have average longer steps than $\mathcal{D}_{pruned}$, but still provide valuable training signals for improving accuracy on complex queries. The training objective is:
$$\mathcal{L}_{hybrid} = -\sum_{\tau^* \in \mathcal{D}_{hybrid}} \log P_{\mathcal{M}}(\tau^*)$$

This strategy allows the model to learn efficient search patterns while retaining the capability to handle complex queries requiring longer but necessary reasoning chains, achieving an optimal trade-off between efficiency and accuracy.

\begin{table*}[t!]
\centering
\small 
\begin{tabular}{l cccc cccc}
\toprule
\multirow{2}{*}{\textbf{Method}} & \multicolumn{4}{c}{\textbf{xbench-deepsearch}} & \multicolumn{4}{c}{\textbf{Browsecomp}} \\
\cmidrule(lr){2-5} \cmidrule(lr){6-9}
& Acc $\uparrow$ & F-AE $\uparrow$ & Rounds $\downarrow$ & Token $\downarrow$ & Acc $\uparrow$ & F-AE $\uparrow$ & Rounds $\downarrow$ & Token $\downarrow$ \\
\midrule
\multicolumn{9}{l}{\textit{Close-sourced System}} \\
OpenAI o3           & 0.670          & -       & -       & -      & 0.497          & -       & -       & -       \\
OpenAI DeepResearch & -              & -       & -       & -      & 0.515          & -       & -       & -       \\
Claude-4-Sonnet     & 0.646          & -       & -       & -      & 0.122          & -       & -       & -       \\
\midrule
\multicolumn{9}{l}{\textit{Open-sourced Agent}} \\
Kimi-K2-Instruct-0905$^*$   & 0.540          & 0.686   & 5.98    & 1316 & 0.094          & 0.169   & 16.65   & 3426          \\
Qwen3-235B-A22B-Instruct-2507$^*$    & 0.490          & 0.637   & 8.84  & 938          & 0.046          & 0.087   & 13.70 & 1837 \\
WebExplorer-8B & 0.517 & 0.659 & 9.05 & 2246 & 0.137 & 0.229 & 29.43 & 6289 \\
\rowcolor{lightblue}
Tongyi-DeepResearch$^*$   & \underline{0.713} & 0.779 & 14.26   & 6918          & 0.410 & 0.385 & 63.70   & 12014         \\
\midrule
\multicolumn{9}{l}{\textit{Pruning Method(vs. Tongyi-DeepResearch)}} \\
\rowcolor{lightblue}
Prompt Control        & 0.676          & 0.763   & {12.50}   & 6321          & 0.373          & 0.372   & 62.80   & 12222         \\
\rowcolor{lightblue}
Coarse Prune       & 0.603          & 0.725   & 8.85   & 4774          & 0.220          & 0.326   & 37.10   & 8365         \\
\rowcolor{lightblue}
WebClipper (Eff)    & \underline{0.713} & \underline{0.792} & {10.81} & {5931} & \underline{0.427} & \textbf{0.431} & {56.50} & {10599} \\
\rowcolor{lightblue}
WebClipper (Hybrid) & \textbf{0.733} & \textbf{0.797} & 12.57   & {6205} & \textbf{0.467} & \underline{0.428} & {60.42} & {11507} \\
\midrule\midrule 
\multirow{2}{*}{\textbf{Method}} & \multicolumn{4}{c}{\textbf{GAIA}} & \multicolumn{4}{c}{\textbf{HLE}} \\
\cmidrule(lr){2-5} \cmidrule(lr){6-9}
& Acc $\uparrow$ & F-AE $\uparrow$ & Rounds $\downarrow$ & Token $\downarrow$ & Acc $\uparrow$ & F-AE $\uparrow$ & Rounds $\downarrow$ & Token $\downarrow$ \\
\midrule
\multicolumn{9}{l}{\textit{Close-sourced System}} \\
OpenAI o3           & -              & -       & -       & -      & 0.249          & -       & -       & -       \\
OpenAI DeepResearch & 0.674          & -       & -       & -      & 0.266          & -       & -       & -       \\
Claude-4-Sonnet     & 0.683          & -       & -       & -      & 0.203          & -       & -       & -       \\
\midrule
\multicolumn{9}{l}{\textit{Open-sourced Agent}} \\
Kimi-K2-Instruct-0905$^*$  & 0.469          & 0.625   & 6.45    & 1281 & 0.146          & 0.253   & 5.17 & 2349 \\
Qwen3-235B-A22B-Instruct-2507$^*$    & 0.456          & 0.612   & 7.14  & 1128          & 0.199          & 0.327   & 7.45 & 2960 \\
WebExplorer-8B & 0.372 & 0.521 & 12.88 & 3560 & 0.116 & 0.203 & 15.52 & 6579 \\
\rowcolor{lightblue}
Tongyi-DeepResearch$^*$   & 0.682 & 0.733 & 20.56   & 7378          & \underline{0.358} & 0.487 & 23.92  & 13664         \\
\midrule
\multicolumn{9}{l}{\textit{Pruning Method (vs. Tongyi-DeepResearch)}} \\
\rowcolor{lightblue}
Prompt Control        & 0.663          & 0.730   & {18.70}   & 6752          & 0.349          & 0.479   & 23.91  & 14107         \\
\rowcolor{lightblue}
Coarse Prune       & 0.514          & 0.638   & 15.60   & 4068          & 0.327          & 0.467   & 18.03  & 11851         \\
\rowcolor{lightblue}
WebClipper (Eff)    & \underline{0.684} & \textbf{0.760} & {14.44} & {4756} & 0.353          & \underline{0.492} & {18.60}  & {11458} \\
\rowcolor{lightblue}
WebClipper (Hybrid) & \textbf{0.695} & \underline{0.744} & 19.92   & {6635} & \textbf{0.361} & \textbf{0.495} & {21.07}  & {13532} \\
\bottomrule
\end{tabular}
\caption{Performance comparison across various web agent benchmarks. The comparison for best (\textbf{bold}) and second-best (\underline{underline}) results is conducted between the base model (Tongyi-DeepResearch) and the Pruning Methods, highlighted in light blue. The $\uparrow$ arrow indicates that higher values are better, while $\downarrow$ indicates lower values are better. $^*$ denotes the result is conducted by ourselves in a unified environment.}
\label{tab:results_all_benchmarks}
\end{table*}

\section{Experiments}
\subsection{Experimental Settings}
\textbf{Evaluation Metrics.} We evaluate web agents from three perspectives:

1) Accuracy: Accuracy (Acc) measured using LLM-as-Judge with o3-mini \cite{o3} as the evaluator.

2) Efficiency: Tool-call rounds and token consumption during inference.

3) F-AE Score: Inspired by the F1 score \cite{f1}, we propose \textbf{F-AE Score} to measure an agent’s ability to balance accuracy and efficiency:
  \[
{\text{F-AE}} = 
  2 \times \frac{\text{Acc} \times \bigl(1 - \frac{\text{Rounds}}{\text{Max\_Rounds}}\bigr)}
           {\text{Acc} + \bigl(1 - \frac{\text{Rounds}}{\text{Max\_Rounds}}\bigr)} ,
  \]
  where \(\text{Max\_Rounds}\) is the maximum number of tool calls allowed in the experiment. Following common practice \cite{tongyidrs}, we set \(\text{Max\_Rounds}=100\). F-AE penalizes both low accuracy and excessive tool usage, thereby avoiding over-optimization of either dimension alone. More explanations can be found in Appendix \ref{F-AE_score}.

\textbf{Datasets.} We conduct evaluations on four widely-used web agent benchmarks: xbench-deepsearch \cite{xbench}, Browsecomp \cite{bc_en}, GAIA \cite{mialon2023gaia}, and HLE \cite{hle}. For GAIA, we use the 103 text-only subset from its development set. For HLE, we follow the setup of previous studies \cite{webthinker} and use a 500 text-only subset.

\textbf{Baselines.}
Our comparison includes both closed-source and open-source agents. Closed-source systems include OpenAI o3 \cite{o3}, OpenAI DeepResearch \cite{openaidr}, and Claude-4-Sonnet \cite{claude4}; test results are cited from their official reports. The open-source agents include Kimi-K2-Instruct-0905 \cite{kimik2},  Qwen3-235B-A22B-Instruct-2507 \cite{yang2025qwen3technicalreport}, WebExplorer \cite{webexplorer} and Tongyi-DeepResearch. As trajectory pruning is underexplored, we design two baselines:
1) \textit{Prompt Control}: We add instructions to the agent's system prompt, explicitly asking it to avoid irrelevant information and repetitive validation, and to control the number of tool calls.
2) \textit{Coarse Prune}: We use Qwen3-235B-A22B-Instruct-2507 to directly identify and remove turns from the trajectory that it deems redundant. The resulting coarsely pruned trajectories are then used for SFT.

\textbf{Implementation.} We use Tongyi-DeepResearch (30B-A3B) \cite{tongyidrs} as the base web agent $\mathcal{M}$. Trajectories are distilled from public QA datasets, including WebShaper, WebDancer, WebExplorer, TaskCraft, and Voyager. We adopt Qwen3-235B-A22B-Instruct-2507 as the extractor and rewriting model for state graph construction and thought rewriting. The extractor and rewriting model is deployed on 8×H800 (80GB) GPUs. Under this setup, the overall extraction and rewriting can be completed within 1 day. Training is conducted on 32×H800 GPUs with a learning rate of 5e-6 and a cosine decay schedule. For WebExplorer, we reproduce its results ourselves. For other open-source models that do not report tool and token usage, we reproduce them by deploying on H800 GPUs within the Tongyi-DeepResearch environment. For web content retrieval, we use the Serper API \cite{serpapi} for search and Jina Reader \cite{jina} for URL parsing. To reduce evaluation variance, each model is run three times with different random seeds, and we report the average Pass@1 and corresponding efficiency metrics.

\begin{figure*}
    \centering
    \includegraphics[width=\textwidth]{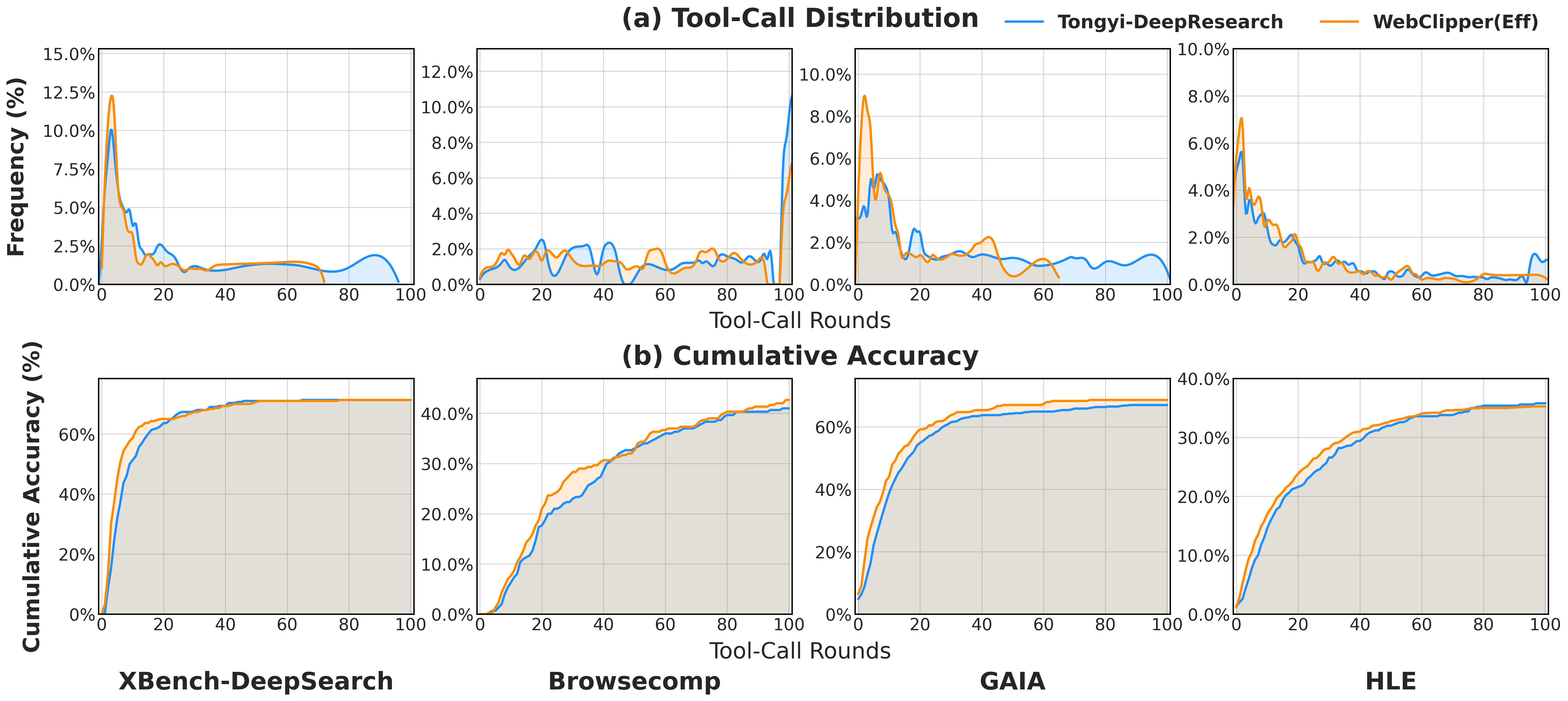}
    \caption{Comparison of tool-call distribution and cumulative accuracy.}
    \label{fig:tool_distribution}
\end{figure*}

\subsection{Main Results}

We organize our experimental investigation around four research questions (RQ): 

\textbf{RQ1}: Is WebClipper an effective pruning strategy? 

\textbf{RQ2}: How does WebClipper compare with direct pruning approaches?

\textbf{RQ3}: How well does the F-AE Score balance the accuracy-efficiency trade-off in web agents? 

\textbf{RQ4}: Are the key components of WebClipper effective?

\begin{table*}[t]
\centering
\resizebox{\textwidth}{!}{
\begin{tabular}{l c>{\columncolor{lightblue}}c cc c>{\columncolor{lightblue}}c cc c>{\columncolor{lightblue}}c cc c>{\columncolor{lightblue}}c cc}
\toprule
& \multicolumn{4}{c}{\textbf{xbench-deepresearch}} & \multicolumn{4}{c}{\textbf{Browsecomp}} & \multicolumn{4}{c}{\textbf{GAIA}} & \multicolumn{4}{c}{\textbf{HLE}} \\
\cmidrule(lr){2-5} \cmidrule(lr){6-9} \cmidrule(lr){10-13} \cmidrule(lr){14-17}
\textbf{Method} & \textbf{Acc} $\uparrow$ & \textbf{F-AE} $\uparrow$ & \textbf{Rounds} $\downarrow$ & \textbf{Token} $\downarrow$ & \textbf{Acc} $\uparrow$ & \textbf{F-AE} $\uparrow$ & \textbf{Rounds} $\downarrow$ & \textbf{Token} $\downarrow$ & \textbf{Acc} $\uparrow$ & \textbf{F-AE} $\uparrow$ & \textbf{Rounds} $\downarrow$ & \textbf{Token} $\downarrow$ & \textbf{Acc} $\uparrow$ & \textbf{F-AE} $\uparrow$ & \textbf{Rounds} $\downarrow$ & \textbf{Token} $\downarrow$ \\
\midrule
Tongyi-DeepResearch & 0.713          & 0.779          & 14.26          & 6918           & 0.410          & 0.385          & 63.70          & 12014          & 0.682          & 0.733          & 20.56          & 7378           & 0.358          & 0.487          & 23.92          & 13664 \\
Unpruned-Distill         & \textbf{0.746} & 0.785          & 17.13          & 6317           & \textbf{0.467} & 0.408          & 63.80          & 11703          & 0.683          & 0.722          & 23.51          & 6992           & \textbf{0.363} & \underline{0.492} & 23.70          & 14099 \\
WebClipper (Eff)    & 0.713          & \underline{0.792} & 10.81          & 5931           & 0.427          & \textbf{0.431} & 56.50          & 10599          & \underline{0.684} & \textbf{0.760} & 14.44          & 4756           & 0.353          & \underline{0.492} & 18.60          & 11458 \\
WebClipper (Hybrid) & \underline{0.733} & \textbf{0.797} & 12.57          & 6205           & \textbf{0.467} & \underline{0.428} & 60.42          & 11507          & \textbf{0.695} & \underline{0.744} & 19.92          & 6635           & \underline{0.361} & \textbf{0.495} & 21.07          & 13532 \\
\bottomrule
\end{tabular}%
}
\caption{Performance comparison of different training strategies.}
\label{tab:data_ablation}
\vspace{-3mm}
\end{table*}

\textbf{Overall Performance (RQ1).} Tables \ref{tab:results_all_benchmarks} present the main results. We highlight several \textbf{key observations: 1)} WebClipper(Eff) achieves leading performance among open-source models while reducing resource consumption. Compared to the Tongyi-DeepResearch baseline, it reduces token usage by \textbf{19.4\%} and tool-call rounds by \textbf{21\%} on average across all benchmarks, while maintaining comparable or even superior accuracy. This demonstrates the effectiveness of efficiency-oriented training in preserving task accuracy while significantly improving search efficiency. \textbf{2)} WebClipper(Hybrid) further improves accuracy with acceptable resource consumption. It achieves the best accuracy among all open-source models, with an average improvement of 4.8\% over the base model, while simultaneously reducing tool-call rounds by 7\%. This validates our hybrid evolution strategy's ability to balance efficiency and accuracy optimization. \textbf{3)}  Further analysis in Figure \ref{fig:tool_distribution} (a) shows that WebClipper(Eff)'s tool-call distribution is concentrated in lower-round buckets compared to the baseline, and Figure \ref{fig:tool_distribution} (b) indicates WebClipper(Eff)'s accuracy curve converges much earlier, indicating superior performance in resource-constrained (low-round) scenarios. These results confirm that WebClipper effectively evolves agents to be more efficient without sacrificing, and sometimes even improving, their information-seeking capabilities.

\textbf{Comparison with Pruning Baselines (RQ2).} Results in Table \ref{tab:results_all_benchmarks} demonstrate WebClipper's superiority over naive pruning strategies: \textbf{1)} Prompt-based pruning is insufficient. Compared to WebClipper(Eff), Prompt Control achieves only a marginal reduction in tool calls while suffering noticeable accuracy degradation. This suggests that directly prompting pre-trained web agents for efficiency is ineffective. \textbf{2)} Coarse-grained pruning causes severe performance drops. The Coarse Prune baseline, which relies on a single LLM to construct training samples through directly identifying redundant rounds, leads to a substantial accuracy drop. This indicates that trajectory optimization requires fine-grained, structured analysis rather than coarse judgment. In contrast, WebClipper's structured, graph-based distillation process allows for precise and reliable identification of redundancies, making it a far more effective pruning strategy.

\textbf{Validity of F-AE Score (RQ3).} The F-AE Score proves to be a balanced metric that avoids bias toward either dimension. As shown in Table \ref{tab:results_all_benchmarks}: \textbf{1)} Despite using shorter rounds, Kimi-K2-Instruct-0905 scores low on F-AE due to their inferior accuracy, preventing the metric from rewarding efficiency alone. \textbf{2)} Although the accuracy of Tongyi-DeepResearch is close to WebClipper(Eff), its longer tool-call rounds result in lower F-AE scores, demonstrating the metric's sensitivity to efficiency. \textbf{3)} WebClipper(Eff) achieves leading F-AE scores by maintaining high accuracy without excessive tool usage, reflecting its superior efficiency-accuracy balance. These patterns show that F-AE does not over-favor either accuracy or efficiency alone; instead, it rewards models that achieve a balanced performance. This supports F-AE as a reasonable and practically useful metric for evaluating web agents. Further explanation can be found in Appendix \ref{F-AE_score}. 

\subsection{Ablation Study}

\begin{figure}
    \centering
    \includegraphics[width=\linewidth]{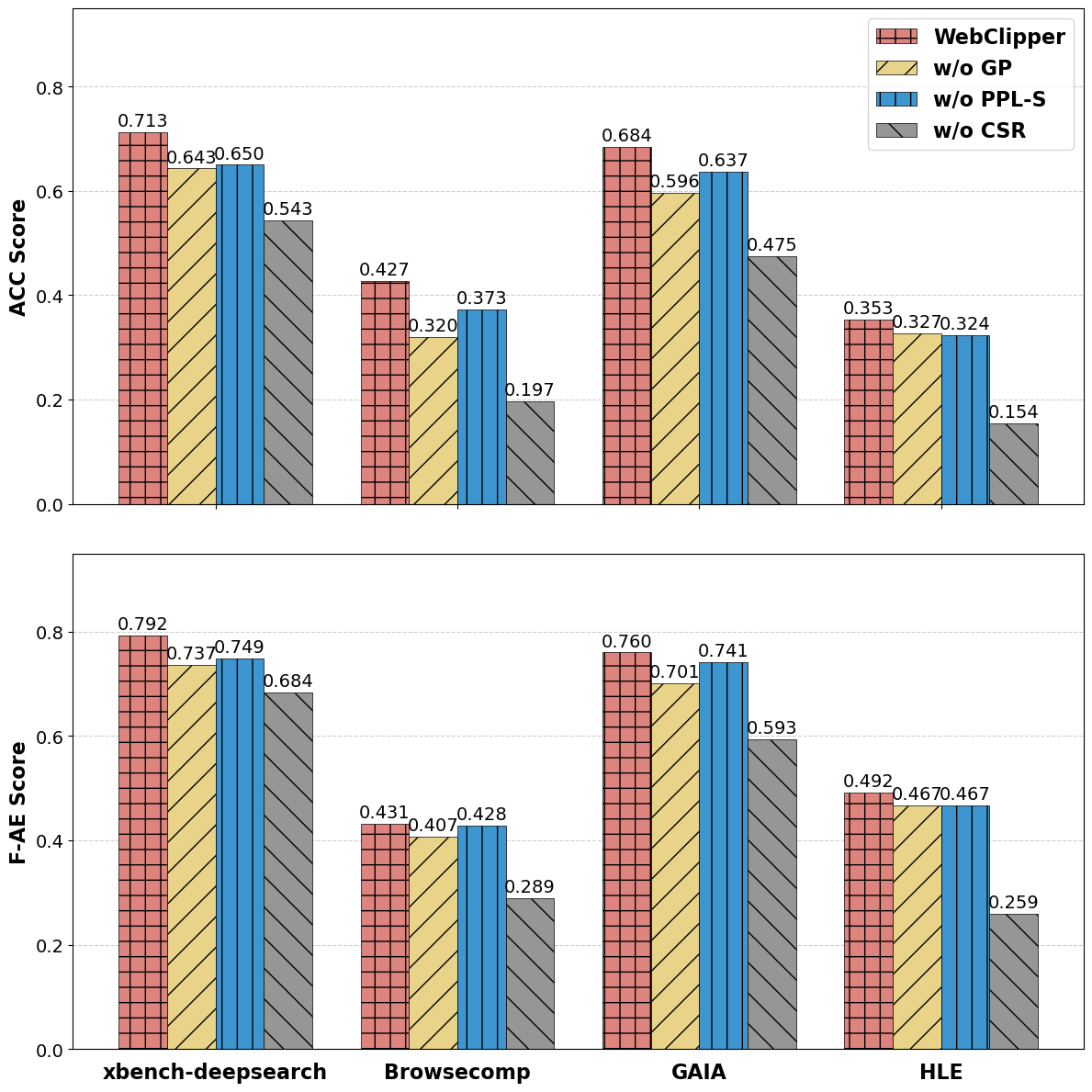}
    \caption{Ablation Study of the key components of WebClipper.}
    \label{fig:components_ablation}
    \vspace{-3mm}
\end{figure}

We now investigate \textbf{RQ4}: Are the key components of WebClipper effective? We conduct ablations on three aspects: the graph-based pruning method, the coherence-aware rewriting strategy, and the agent evolution strategy.

\textbf{Ablation on Pruning Method \& Rewriting Strategy.}
We evaluate three variants: (1) \textit{w/o GP}, replacing graph-based pruning with Coarse Prune but retaining the rewriting strategy; (2) \textit{w/o PPL-S}, removing PPL-based selection and using the first generated rewriting as the final thought in trajectories; (3) \textit{w/o CSR}, replacing context-aware selective rewriting with unconditional rewriting of all thoughts without providing the historical context. As shown in Figure \ref{fig:components_ablation}, removing any component causes performance degradation. The decline of \textit{w/o GP} can be attributed to the fact that single-pass LLM comprehension struggles with long trajectories. The drop in \textit{w/o PPL-S} validates PPL-based filtering in maintaining alignment with the base model's reasoning style. Most critically, \textit{w/o CSR} leads to catastrophic collapse, confirming that naive rewriting without understanding context breaks reasoning coherence.

\textbf{Ablation on Evolution Strategy.} 
We compare three training strategies: WebClipper(Eff), WebClipper(Hybrid), and ``Unpruned-Distill''. ``Unpruned-Distill'' follows the commonly adopted self-evolve paradigm \cite{aksitov2023rest}, where original unpruned trajectories with \(0 < \mathrm{PR}(q) \le 0.5\) are directly used for SFT, and data is directly obtained from  Section~\ref{rejection_sample}. As shown in Table~\ref{tab:data_ablation}, Unpruned-Distill improves accuracy over the base model but increases tool-call rounds, amplifying both strengths and inefficiencies. WebClipper(Eff) achieves the lowest resource usage while maintaining accuracy comparable to the base model, making it preferable when efficiency is the primary concern. WebClipper(Hybrid) provides a more balanced option: relative to both Unpruned-Distill and the base model, it uses fewer rounds, attains accuracy clearly above the base model and close to Unpruned-Distill, and achieves the better F-AE scores. In practice, WebClipper(Eff) suits cost-sensitive deployments, whereas WebClipper(Hybrid) delivers a more comprehensive improvement in both efficiency and accuracy.

\subsection{Analysis and Discussion}
Beyond efficiency gains, WebClipper also improves accuracy. We attribute this to the reasoning patterns induced by our pruned data, which trains the agent to focus on critical-path information. Existing web agents often fall into failure modes where they become stuck in unproductive branches, drift from the core objective, or enter cyclic reasoning loops. As shown in our case studies in Appendix \ref{case_study}, over-focusing on trivial details can make the agent lose sight of the main goal. This not only reduces efficiency but also harms accuracy by inflating context length: an overly long context increases the risk that useful clues are drowned out by a mass of irrelevant, more recent tool interactions. Our pruning method counteracts this by constructing training samples in which irrelevant or repetitive tool-calling rounds are removed.

We also find that WebClipper’s efficiency gains are particularly notable on the GAIA dataset, where tool-call rounds are reduced by about 30\%. We attribute this to the dataset’s characteristics: around 15\% of its questions are brain teasers or logical puzzles that rely on the model’s intrinsic abstract reasoning and instruction-following, rather than long-horizon tool use. Excessive emphasis on external tools during training can not improve performance on such problems. Our method prevents the model from over-relying on external tools in these cases, substantially reducing unnecessary tool calls.

\section{Conclusion}
In this paper, we propose WebClipper, an innovative trajectory pruning method for web agents. We model web agent trajectories as state graphs and perform pruning on them. We further introduce two agent evolution strategies, which significantly reduce the number of tool calls while maintaining or even improving the agent’s accuracy. In addition, we propose the F-AE score to better evaluate the overall capability of web agents in terms of both accuracy and efficiency. Extensive experiments demonstrate that WebClipper is an effective approach for balancing accuracy and efficiency.

\section*{Limitations}

WebClipper has achieved significant improvements in the efficiency of web agents, but there remain several limitations that point to future directions. 

First, WebClipper inherits the planning and reasoning capabilities of the base model it distills from—if the base model's performance is poor, the pruning process can only remove redundancy within those suboptimal trajectories rather than fundamentally improving the search strategy. Future work could explore integrating WebClipper with reinforcement learning or online learning mechanisms to enable the agent to discover novel, more efficient search patterns beyond those present in the base model's behavior. Second, our pruning method is trained and evaluated on trajectories from specific web agent benchmarks that primarily involve search, web browsing, and code execution, leaving the generalization to emerging tool types (e.g., multimodal tools, database queries, or API integrations) unexplored. Extending WebClipper's graph-based framework to accommodate diverse action spaces and information modalities represents a valuable direction for building more versatile and efficient agents across broader application domains.

\bibliography{custom}

\clearpage

\appendix

\section{Design of F-AE Score}
\label{F-AE_score}
In this section, we provide a more detailed explanation of the F-AE Score, drawing an analogy to the classic F1-Score in information retrieval, and clarifying the design choices behind its formulation.

\subsection{Background: The F1 Score}
The F1 score is a widely-used metric in classification tasks that harmonizes precision and recall through their harmonic mean:
$$\text{F1} = \frac{2 \times \text{Precision} \times \text{Recall}}{\text{Precision} + \text{Recall}}$$

The key insight of F1 is that it balances two competing objectives—precision (quality of positive predictions) and recall (coverage of actual positives)—in a way that penalizes extreme imbalance. Unlike the arithmetic mean, which would be $\frac{\text{Precision} + \text{Recall}}{2}$, the harmonic mean is more sensitive to low values. For instance, if Precision = 1.0 but Recall = 0.1, the arithmetic mean yields 0.55, while F1 yields only 0.18, reflecting that a model excelling in only one dimension is suboptimal.

\subsection{Motivation for F-AE Score}

When evaluating Web Agents, we face an analogous trade-off between two competing objectives:
\begin{itemize}
    \item Accuracy (\(\text{Acc}\)): how often the agent produces a correct answer.
    \item Efficiency (\(E\)): how economical the agent is in its use of tool-calling rounds.
\end{itemize}

Existing evaluation paradigms often optimize these metrics in isolation. To holistically assess agent quality, we need a metric that captures their joint optimization.

\subsection{Design of F-AE Score}
Our F-AE Score follows exactly the same philosophy as F1 Score, but replaces precision/recall with Acc and (\(E\)):

We first normalize the number of tool-calling rounds to an ``efficiency score'' in \([0,1]\):
\[
E = 1 - \frac{\text{Rounds}}{\text{Max\_Rounds}},
\]
where \(\text{Rounds}\) is the average number of tool-call turns used by the agent, and \(\text{Max\_Rounds}\) is the maximum allowable rounds in the deployment scenario (set to 100 in our experiments). Intuitively, if an agent uses \(\text{Rounds} = \text{Max\_Rounds}\), then \(E = 0\), i.e., ``maximally inefficient''; if an agent uses very few rounds, say \(\text{Rounds} \approx 0\), then \(E \approx 1\), i.e., ``highly efficient''.

We then define F-AE Score as the harmonic mean of accuracy and efficiency:
\[
\text{F-AE} = 2 \times \frac{\text{Acc} \times E}{\text{Acc} + E}
= 
2 \times \frac{\text{Acc} \times \bigl(1 - \frac{\text{Rounds}}{\text{Max\_Rounds}}\bigr)}
           {\text{Acc} + \bigl(1 - \frac{\text{Rounds}}{\text{Max\_Rounds}}\bigr)}
\]
where both Acc and \(E\) are normalized to $[0, 1]$, ensuring F-AE $\in [0, 1]$ for interpretability. A higher F-AE means better overall performance, taking both dimensions into account. This makes it easy to compare different Web Agents or training strategies.

Using the harmonic mean between \(\text{Acc}\) and \(E\) has several desirable properties:

1. \textbf{Balance between accuracy and efficiency}. If either accuracy or efficiency is low, F-AE will be low. For example:
\begin{itemize}
    \item A model with high accuracy but extremely long trajectories (\(E \approx 0\)) will receive a low F-AE. 
    \item A model with very short trajectories but poor accuracy (\(\text{Acc} \approx 0\)) will also receive a low F-AE.  
   This matches our intuitive requirement that a ``good'' Web Agent must be both effective and efficient.
\end{itemize}

2.\textbf{No arbitrary dominance of one dimension.}  
Unlike a simple weighted sum (e.g., \(\alpha \cdot \text{Acc} + (1-\alpha) \cdot E\)), the harmonic mean is far less tolerant of one dimension being much smaller than the other. This prevents scenarios where:
\begin{itemize}
    \item Slight gains in accuracy justify arbitrarily large increases in rounds
    \item Slight savings in rounds justify large accuracy drops.
\end{itemize}
In other words, F-AE inherently discourages extreme trade-offs.

\subsection{Effect of Max\_Rounds and Scaling}

The parameter \(\text{Max\_Rounds}\) controls how aggressively we penalize tool usage:

\[
E = 1 - \frac{\text{Rounds}}{\text{Max\_Rounds}}.
\]

When \(\text{Rounds} \ll \text{Max\_Rounds}\), \(E\) is close to 1, so efficiency is considered good and F-AE is mainly determined by accuracy.  
When \(\text{Rounds}\) approaches \(\text{Max\_Rounds}\), \(E\) decreases toward 0, pulling F-AE down even if accuracy remains high.

In our experiments, \(\text{Max\_Rounds} = 100\) is chosen to reflect the typical upper bound used in Deep Research–style Web Agents. In principle, \(\text{Max\_Rounds}\) can be adjusted to match different deployment constraints (e.g., stricter limits in latency-critical settings).

An important point is that F-AE is relative to the chosen budget: if all methods are evaluated with the same \(\text{Max\_Rounds}\), F-AE provides a fair way to compare them under that shared resource regime.

\section{Implementation Details}
\label{sec:method_details}
This appendix elaborates on the implementation of our trajectory pruning and rewriting pipeline, providing conceptual descriptions and the specific prompts used.
\subsection{Details of Rejection Sampling}
We use the public QA datasets to distill trajectories. The used dataset includes WebDancer (200 samples), WebShaper (500 samples), WebExplorer (100 samples), Voyager (a subset consisting of 5k samples), and TaskCraft (a subset consisting of 4k samples). In the Tongyi-DeepResearch environment, we ran all samples four times, keeping those with a pass rate $0 < \mathrm{PR}(q) \le 0.5,$. This data was then used for subsequent pruning.

\subsection{State Graph Construction}
The construction of the state graph \(\mathcal{G}\) from a raw trajectory \(\tau\) is a two-phase process orchestrated by an LLM extractor.

\begin{figure*}
    \centering
    \includegraphics[width=\linewidth]{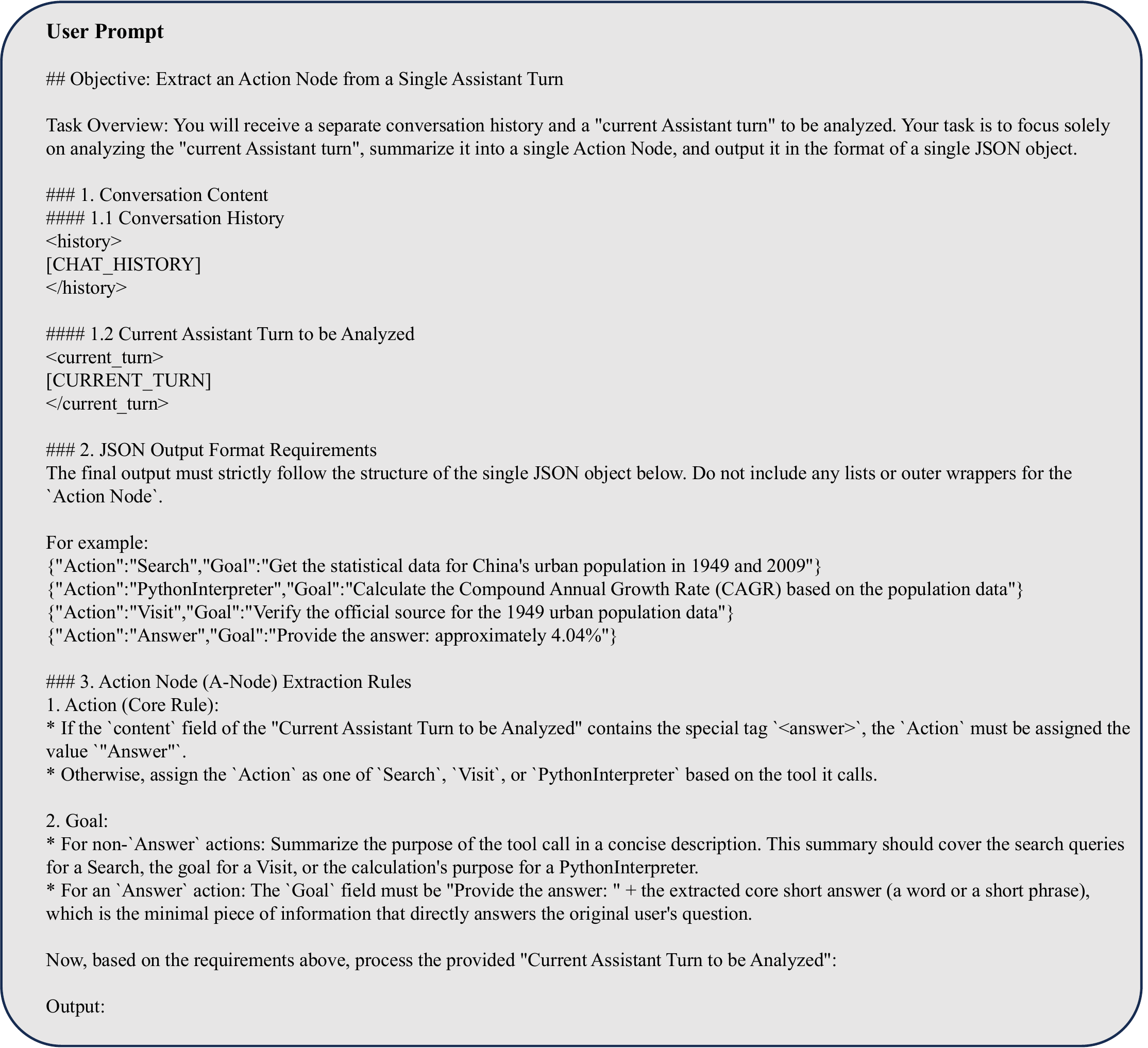}
    \caption{The Prompt of Action Node Extraction}
    \label{fig:action_node_prompt}
\end{figure*}

\paragraph{Phase 1: Action Node Extraction}
First, we process the trajectory to identify each assistant turn uniquely. Each turn, consisting of a thought-action pair $(t_k, a_k)$, is mapped to a corresponding Action Node $A_k$. We employ an LLM extractor (example prompt is shown in Figure \ref{fig:action_node_prompt}) that receives the conversational history up to step $k-1$ and the current turn's content. The extractor's task is to summarize this turn into a compact JSON object with two fields: an ``Action'' type (e.g., Search, PythonInterpreter, Answer) and a ``Goal'' description. This process is parallelized across all turns in the trajectory for efficiency, yielding the complete set of action vertices, \(\mathcal{V}^A\).

\paragraph{Phase 2: Iterative Information and Edge Construction}
With the action nodes \(\mathcal{V}^A\) established, we iteratively build the information nodes \(\mathcal{V}^I\) and the dependency edges \(\mathcal{E}\). The process is initialized with a graph containing only the initial query node \(I_0\) and the first action node \(A_1\), connected by an edge $(I_0, A_1)$.

\begin{figure*}
    \centering
    \includegraphics[width=\linewidth]{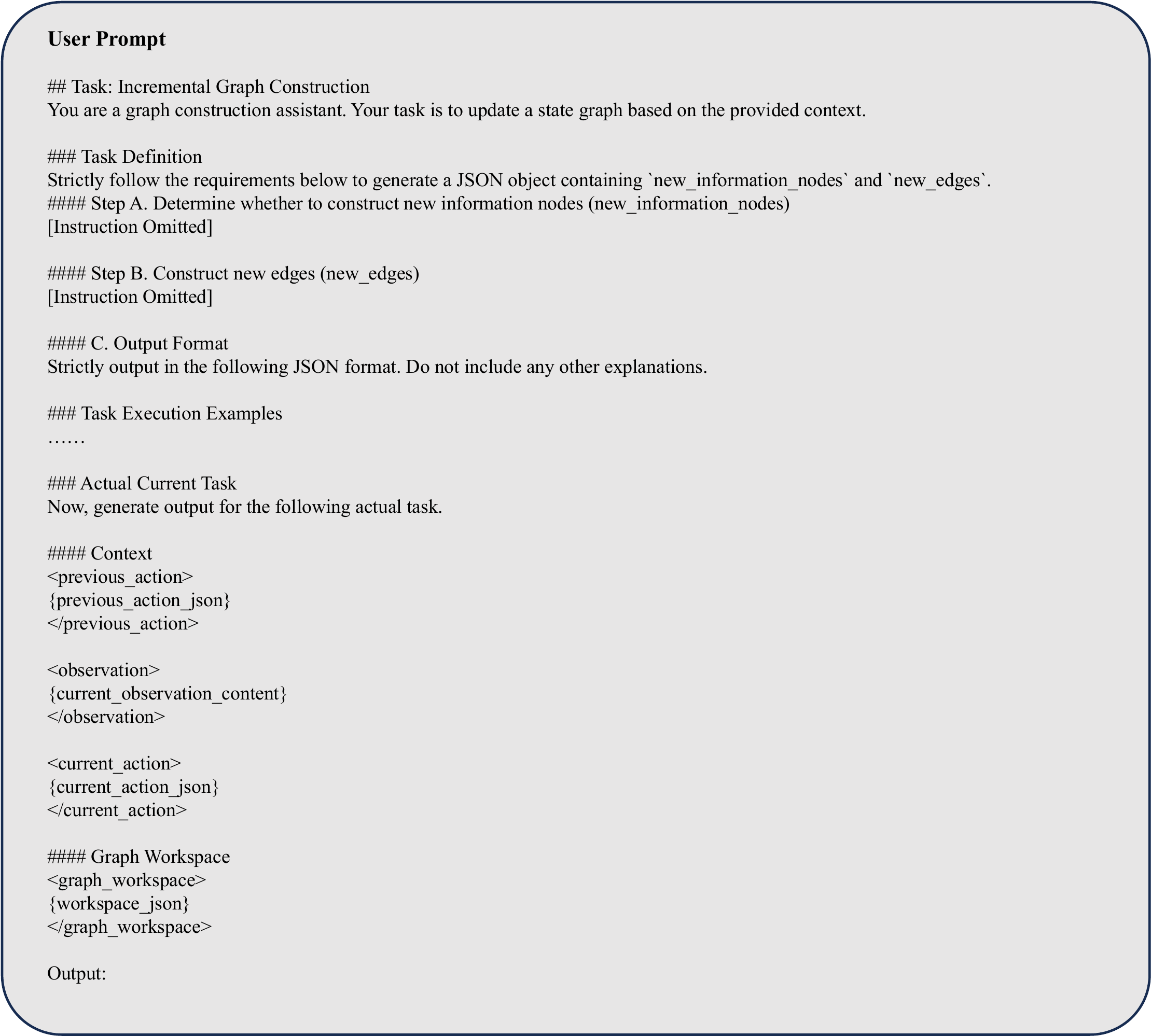}
    \caption{The Prompt of Iterative Information and Edge Construction}
    \label{fig:info_edge_prompt}
\end{figure*}

We then iterate from $k=1$ to $T-1$. In each iteration, the LLM extractor is prompted (example prompt is shown in Figure \ref{fig:info_edge_prompt}) with the current graph state and a snippet of the trajectory: $(A_k, o_k, A_{k+1})$, where $o_k$ is the observation received after action $A_k$. The LLM performs two functions:
\begin{enumerate}
    \item \textbf{Decomposing Observations:} It analyzes $o_k$ to extract atomic units of information. For each unit, it checks for semantic equivalence with existing nodes in \(\mathcal{V}^I\). If a match is found, an edge $A_k \rightarrow I_{\text{existing}}$ is added. Otherwise, a new information node $I_{\text{new}}$ is created and added to \(\mathcal{V}^I\), along with an edge $A_k \rightarrow I_{\text{new}}$.
    \item \textbf{Linking Actions:} It analyzes $A_{k+1}$ to identify which information nodes in the current graph (including any newly created ones) served as its basis. For each identified supporting node $I'$, an edge $I' \rightarrow A_{k+1}$ is added.
\end{enumerate}
This iterative process continues until all actions and observations have been incorporated, resulting in the final state graph \(\mathcal{G}\).

\subsection{Pruning via MNDAG and Majority Vote}

\paragraph{MNDAG Identification}
Given the state graph \(\mathcal{G}\), we identify the minimal necessary subgraph using a two-stage algorithm, detailed in Algorithm \ref{alg:mndag}. This algorithm approximates the Minimal-cost Necessary Directed Acyclic Graph (MNDAG).

\begin{algorithm}[h!]
\caption{MNDAG Identification Algorithm}
\label{alg:mndag}
\begin{algorithmic}[1]
\State \textbf{Input:} State Graph $\mathcal{G}=(\mathcal{V}, \mathcal{E})$, source $I_0$, sink $A_T$
\State \textbf{Output:} Set of necessary action nodes $\mathcal{A}^\star$

\Statex \textit{// Step 1: Forward search for shortest path costs}
\State Define cost function: $c(v) = 1$ if $v \in \mathcal{V}^A$, else $c(v) = 0$.
\State Run Dijkstra's algorithm starting from $I_0$ to compute the shortest distance $d(v)$ and predecessor $p(v)$ for every node $v \in \mathcal{V}$.

\Statex \textit{// Step 2: Backward traversal to identify necessary nodes}
\State Initialize necessary node set $\mathcal{V}^\star \leftarrow \{A_T\}$.
\State Initialize a queue for traversal $Q \leftarrow [A_T]$.
\While{$Q$ is not empty}
    \State Dequeue a node $v$.
    \If{$v \in \mathcal{V}^A$} \Comment{If node is an Action}
        \For{each predecessor $u$ of $v$ in $\mathcal{G}$}
            \If{$u \notin \mathcal{V}^\star$}
                \State Enqueue $u$ and add to $\mathcal{V}^\star$.
            \EndIf
        \EndFor
    \ElsIf{$v \in \mathcal{V}^I$ and $p(v)$ exists} \Comment{If node is Information}
        \State Let $u \leftarrow p(v)$ \Comment{Get predecessor from Dijkstra's path tree}
        \If{$u \notin \mathcal{V}^\star$}
            \State Enqueue $u$ and add to $\mathcal{V}^\star$.
        \EndIf
    \EndIf
\EndWhile

\Statex \textit{// Step 3: Extract final action set}
\State $\mathcal{A}^\star \leftarrow \{A \mid A \in \mathcal{V}^\star \cap \mathcal{V}^A\}$
\State \textbf{return} $\mathcal{A}^\star$
\end{algorithmic}
\end{algorithm}

\paragraph{Robustness via Majority Vote}
A single LLM-driven graph construction can be prone to inconsistencies. To enhance robustness, we repeat the entire process—from graph construction to MNDAG mining—three times for the same trajectory. This yields three candidate sets of necessary actions: $\mathcal{A}^{\star(1)}, \mathcal{A}^{\star(2)}, \mathcal{A}^{\star(3)}$. A final set $\mathcal{A}^{\star}_{\text{final}}$ is accepted only if at least two of the three candidate sets are identical. If no majority is reached, the pruning for that trajectory is considered unreliable and is discarded, ensuring we only proceed with high-confidence results.

\subsection{Coherence-aware Thought Rewriting}

Simply deleting steps can create logical gaps. When intermediate rounds are pruned, the remaining reasoning steps may still refer to observations that have already been removed, thereby creating incoherent training signals. Consider the following example.

\paragraph{Before pruning (Rounds 1 $\rightarrow$ 2 $\rightarrow$ 3 $\rightarrow$ 4).}
\begin{itemize}
    \item \textbf{Round 1:} Search for ``Nobel Prize in Physics 2020.''  
    \textit{Observation:} The search engine returns relevant results.
    
    \item \textbf{Round 2:} Visit a tangentially related page about the 2019 Nobel Prize.  
    \textit{Observation:} The page contains information about the 2019 laureates, which turns out to be an unproductive branch.
    
    \item \textbf{Round 3 Thought:} \textit{``The previous page was about the 2019 prize, which is not what I need. Let me go back and focus on the 2020 laureates instead.''}
    
    \item \textbf{Round 4:} Visit the correct page for the 2020 Nobel Prize.  
    \textit{Observation:} The page returns the correct answer.
\end{itemize}

\paragraph{After pruning Round 2 (Rounds 1 $\rightarrow$ 3 $\rightarrow$ 4).}
Round 3 still contains the thought:
\begin{quote}
\textit{``The previous page was about the 2019 prize, which is not what I need.''}
\end{quote}
However, the visit to the 2019 page in Round 2 has already been removed. If this pruned trajectory is used directly for training, the model may learn to hallucinate a failed attempt that never occurred in the retained context. As a result, the model could be encouraged to fabricate non-existent observations.

\paragraph{After rewriting.}
The thought in Round 3 can be revised as follows:
\begin{quote}
\textit{``I found relevant search results. Let me visit the page for the 2020 Nobel Prize to get the details.''}
\end{quote}

This rewriting step ensures that the pruned trajectory remains logically self-consistent and free from hallucinated references, making it safer and more effective as training data.

\begin{table*}[t!]
\centering
\resizebox{\textwidth}{!}{
\begin{tabular}{lcccccccccccc}
\toprule
\multirow{2}{*}{\textbf{Method}} 
& \multicolumn{3}{c}{\textbf{xbench-deepsearch}} 
& \multicolumn{3}{c}{\textbf{GAIA}} 
& \multicolumn{3}{c}{\textbf{Browsecomp}} 
& \multicolumn{3}{c}{\textbf{HLE}} \\
\cmidrule(lr){2-4}\cmidrule(lr){5-7}\cmidrule(lr){8-10}\cmidrule(lr){11-13}
& \textbf{Acc}~$\uparrow$ & \textbf{F-AE}~$\uparrow$ & \textbf{Rounds}~$\downarrow$ 
& \textbf{Acc}~$\uparrow$ & \textbf{F-AE}~$\uparrow$ & \textbf{Rounds}~$\downarrow$ 
& \textbf{Acc}~$\uparrow$ & \textbf{F-AE}~$\uparrow$ & \textbf{Rounds}~$\downarrow$ 
& \textbf{Acc}~$\uparrow$ & \textbf{F-AE}~$\uparrow$ & \textbf{Rounds}~$\downarrow$ \\
\midrule
WebExplorer & 0.517 & 0.659 & 9.05 & 0.372 & 0.512 & 12.88 & 0.137 & 0.229 & 29.43 & 0.116 & 0.203 & 15.52 \\
+WebClipper(Eff) & 0.560 & 0.696 & 7.82 & 0.407 & 0.564 & 8.22 & 0.135 & 0.230 & 22.95 & 0.146 & 0.252 & 7.4 \\
\bottomrule
\end{tabular}
}
\caption{Performance of WebClipper based on WebExplorer-8B.}
\label{webexplorer}
\end{table*}

\begin{figure*}
    \centering
    \includegraphics[width=\linewidth]{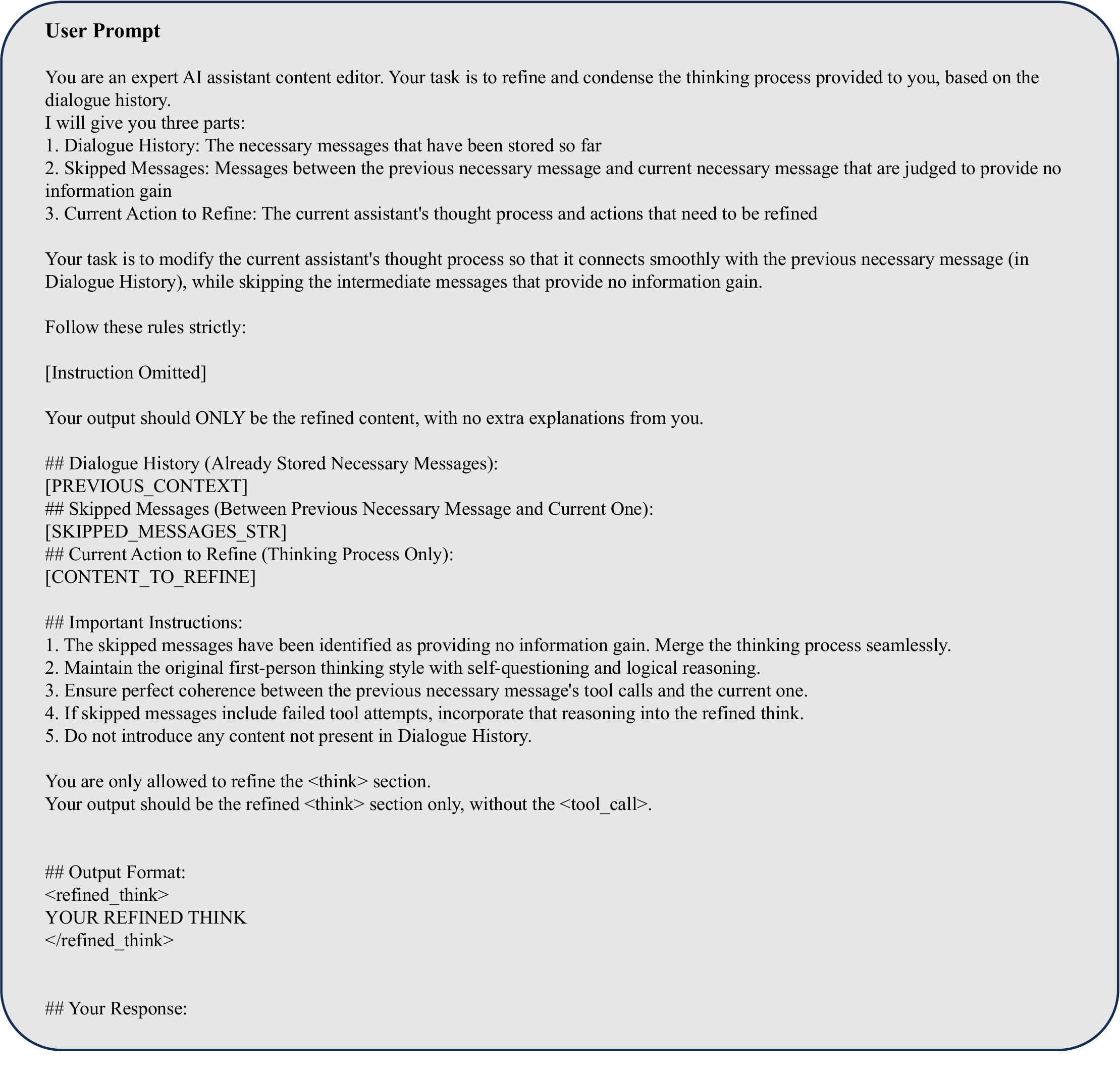}
    \caption{The Prompt of Message Refine}
    \label{fig:message_refine_prompt}
\end{figure*}

\paragraph{Context-aware Selective Rewriting}
We only rewrite thoughts that become disconnected from their new predecessors after pruning. A thought $t^{\text{new}}_{k+1}$ is rewritten if the action $a^{\text{new}}_{k}$ preceding it in the pruned trajectory was not its direct predecessor in the original trajectory. The rewriting LLM (example prompt is shown in Figure \ref{fig:message_refine_prompt}) is conditioned on a comprehensive context:
\begin{itemize}
    \item \textbf{Dialogue History}: The sequence of necessary messages generated so far.
    \item \textbf{Skipped Messages}: The raw content of all intermediate steps that were pruned between the new adjacent steps. This provides the LLM with the knowledge of what occurred in the gap.
    \item \textbf{Current Action to Refine}: The original thought from the step being rewritten.
\end{itemize}
This rich context enables the LLM to generate a new thought that smoothly bridges the logical gap while avoiding hallucinations by not referencing pruned observations directly.

\paragraph{Perplexity-based Selection}
To ensure the rewritten thought aligns with the base model's intrinsic reasoning style, we generate three candidate rewrites for each required modification. We then use the base model itself to calculate the perplexity (PPL) of each candidate. The PPL is computed over the rewritten thought, conditioned on the preceding dialogue history. The candidate with the lowest PPL is selected, which ensures maximal fluency and stylistic consistency with the model's own text distribution.

\section{Additional Experiment Results}
\subsection{Training based on WebExplorer-8B}
To further demonstrate the generalizability of our method, we use the trajectories generated by our WebClipper framework to fine-tune a different and existing web-agent, WebExplorer-8B.

As shown in Table \ref{webexplorer}, our method can perform well on WebExplorer-8B, maintaining model performance while reducing tool-call rounds. This demonstrates that our method is not model-specific and possesses generalization ability.

\subsection{Case Study}
\label{case_study}
\subsubsection{Case 1}
As shown in Table \ref{tab:case_study_1}, the task requires identifying the nano-compound studied in a specific 2012 Scientific Reports article that does not mention ``plasmons'' or ``plasmonics''. The initial agent correctly identifies the target article, ``Diamond photonic crystal slab: Leaky modes and modified photoluminescence emission of surface-deposited quantum dots''. However, it then engages in ``excessively divergent exploration''. It shifts its focus from the primary subject of the paper (the diamond slab) to a trivial detail—the ``surface-deposited quantum dots'' mentioned in the title. This leads to a long chain of tool calls (10+ rounds) to identify the quantum dots' material (``silicon nanocrystals'') and repeatedly validate the initial conditions. This over-exploration, which significantly inflates the context length, causes the model to \textbf{lose sight of the core objective}, mistaking the experimental probe for the main subject of study. In contrast, WebClipper demonstrates a more pruned and effective reasoning path. After identifying the correct article in just two rounds, it directly infers the main subject, ``diamond'', from the title and a concise tool-provided summary. By avoiding the irrelevant deep-dive into the quantum dot material, it prevents context dilution and the risk of forgetting critical initial information. 

\subsubsection{Case 2}

We further demonstrate WebClipper's efficiency gains with a second case (Table \ref{tab:case_study_2}). The task is to calculate the time for Eliud Kipchoge to run to the Moon's perigee. This requires finding two constants (Kipchoge's pace and the Moon's minimum perigee) and performing a calculation. The initial model exhibits clear hallmarks of an unpruned, exhaustive search. It gets bogged down by the slight ambiguity of the term ``record-making marathon pace,'' which could refer to several of Kipchoge's historic runs. This uncertainty triggers multiple, redundant search and visit cycles to verify and re-verify his latest record time, as well as the Moon's perigee distance. Furthermore, it engages in superfluous exploration by calculating the final answer for three different marathon times, even though the question implies a single, definitive pace. This inefficient, cyclical trajectory significantly increases the number of tool calls (over 15 rounds). In contrast, WebClipper adopts a more decisive and linear strategy. It makes a reasonable initial assumption for the record time and proceeds along a direct path: find the two required constants, then compute the result. This pruned approach, consisting of just four rounds, entirely avoids the redundant validation loops and superfluous computations seen in the baseline. This case demonstrates that our training method teaches the model to commit to a reasonable and efficient path, improving performance by eliminating unnecessary and costly over-exploration.

\begin{table*}[h!]
\centering
\caption{Case 1 comparison between WebClipper and Tongyi-DeepResearch.}
\label{tab:case_study_1}
\resizebox{\textwidth}{!}{%
\begin{tabular}{|p{0.5\textwidth}|p{0.5\textwidth}|}
\hline
\multicolumn{2}{|c|}{\cellcolor{gray!20}\parbox{\dimexpr\textwidth-2\tabcolsep\relax}{\centering\textbf{Query:} In Nature journal's Scientific Reports conference proceedings from 2012, in the article that did not mention plasmons'' or plasmonics'', what nano-compound is studied? Don't use the prefix nano in your answer if there is one.}} \\
\hline
\textbf{WebClipper} & \textbf{Baseline Model} \\
\hline
\textit{1. Assistant calls ``search(query=["Scientific Reports 2012 conference..."])}'' & \textit{1. Assistant calls ``search(query=["Scientific Reports 2012 conference..."])}'' \\
\hline
\textbf{Tool Response:} Provides a link to the 2012 conference proceedings page on nature.com.
[Conference Proceedings in 2012 | Scientific Reports](url:...) ... & \textbf{Tool Response:} Provides a link to the 2012 conference proceedings page on nature.com.
[Conference Proceedings in 2012 | Scientific Reports](url:...) ... \\
\hline
\textit{2. Assistant calls ``visit(url=["..."], goal="List articles... and identify which one does not mention plasmons...")''} & \textit{2. Assistant calls ``visit(url=["..."], goal="List conference proceeding articles...")''} \\
\hline
\textbf{Tool Response:} The 2012 conference proceedings... contain four articles. Three articles explicitly mention plasmon-related terms... The article that does not mention plasmons... is ``Diamond photonic crystal slab...''. The studied nano-compound in this article is diamond. & \textbf{Tool Response:} The webpage lists four conference proceeding articles: ``Diamond photonic crystal slab...'', ``2D quasiperiodic plasmonic crystals'', ``Layered plasmonic cloaks...'', etc. \\
\hline
\textit{3. Assistant concludes and provides the final answer.}
``<answer>diamond</answer>'' & \textit{3. Assistant calls ``visit(url=[".../srep00914"], goal="...find the nano-compound studied...")''} \\
\hline
& \textbf{Tool Response:} ...The research investigates the leaky modes of this diamond PhC slab and their effect on the photoluminescence (PL) emission of surface-deposited quantum dots. The specific nano-compound studied is silicon nanocrystals (SiNCs)... \\
\hline
& \textit{4-10. A series of subsequent ``visit'' and ``search'' calls.} The agent repeatedly visits ``srep00914'' to confirm the absence of ``plasmon,'' visits other articles (e.g., ``srep00915'') to confirm the presence of ``plasmon,'' and accesses the PDF version of ``srep00914'' for deeper analysis. \\
\hline
& \textbf{Tool Response:} Responses confirm that ``srep00914'' does not mention ``plasmon'', while other articles do. This reinforces the agent's focus on ``srep00914'' and its contents, particularly the SiNCs. \\
\hline
& \textit{11. Assistant concludes and provides the final answer.}
``<answer>...the nano-compound studied in the only Scientific Reports conference proceeding article from 2012 that does not mention plasmons or plasmonics is silicon nanocrystals...</answer>'' \\
\hline
\end{tabular}%
}
\end{table*}

\begin{table*}[h!]
\centering
\caption{Case 2 comparison between WebClipper and Tongyi-DeepResearch.}
\label{tab:case_study_2}
\resizebox{\textwidth}{!}{%
\begin{tabular}{|p{0.5\textwidth}|p{0.5\textwidth}|}
\hline
\multicolumn{2}{|c|}{\cellcolor{gray!20}\parbox{\dimexpr\textwidth-2\tabcolsep\relax}{\centering\textbf{Query:} If Eliud Kipchoge could maintain his record-making marathon pace indefinitely, how many thousand hours would it take him to run the distance between the Earth and the Moon its closest approach? Please use the minimum perigee value on the Wikipedia page for the Moon when carrying out your calculation. Round your result to the nearest 1000 hours and do not use any comma separators if necessary.}} \\
\hline
\textbf{WebClipper} & \textbf{Baseline Model} \\
\hline
\textit{1. Assistant calls ``PythonInterpreter'' to calculate speed based on a known record (2h 1m 39s).} & \textit{1. Assistant calls ``search(query=["Eliud Kipchoge record making..."])}'' \\
\hline
\textbf{Tool Response:} ``stdout: 20.811...'' & \textbf{Tool Response:} Provides multiple results, including his 2:01:39 record, his sub-2-hour 1:59:40 attempt, and his more recent 2:01:09 record, creating ambiguity. \\
\hline
\textit{2. Assistant calls ``search(query=["Moon perigee minimum distance..."])}'' & \textit{2. Assistant calls ``search(query=["Moon perigee closest approach..."])}'' \\
\hline
\textbf{Tool Response:} Provides multiple values for the Moon's perigee. & \textbf{Tool Response:} Provides multiple values for the Moon's perigee. \\
\hline
\textit{3. Assistant calls ``visit'' on the Wikipedia page for the Moon to find the precise minimum perigee.} & \textit{3. Assistant calls ``visit'' on the Wikipedia page to find the precise minimum perigee.} \\
\hline
\textbf{Tool Response:} ...the minimum perigee distance value listed is 356,400 kilometres. & \textbf{Tool Response:} ...minimum perigee distance... is 356,400 km... \\
\hline
\textit{4. Assistant combines the constants and provides the final answer.}
``<answer>...it would take approximately 17 thousand hours.</answer>'' & \textit{4. Assistant calls ``search'' again to re-verify the latest world record.} \\
\hline
& \textbf{Tool Response:} Confirms the 2022 record is 2:01:09. \\
\hline
& \textit{5-8. A series of ``PythonInterpreter'' calls.} The agent calculates the speed for the 2:01:09 record and then the total time. \\
\hline
& \textit{9-14. A series of subsequent ``search'', ``visit'', and ``PythonInterpreter'' calls.} The agent engages in redundant validation (re-visiting Wikipedia) and superfluous exploration (calculating the final time for the other two marathon paces, 1:59:40 and 2:01:39). \\
\hline
& \textbf{Tool Response:} The calculations for all three marathon paces round to the same final answer. \\
\hline
& \textit{15. Assistant compiles all findings and provides the final answer.}
``<answer>...Eliud Kipchoge would require approximately 17000 thousand hours...</answer>'' \\
\hline
\end{tabular}%
}
\end{table*}

\end{document}